# Analysis of the Fed's communication by using textual entailment model of Zero-Shot classification

Yasuhiro Nakayama[1], Tomochika Sawaki [2]

[1] Mizuho Research & Technologies, Ltd.
[2] Mizuho Bank, Ltd.
yasuhiro.nakayama@mizuho-rt.co.jp, tomochika.sawaki@mizuho-bk.co.jp

**Abstract:** In this study, we analyze documents published by central banks using text mining techniques and propose a method to evaluate the policy tone of central banks. Since the monetary policies of major central banks have a broad impact on financial market trends, the pricing of risky assets, and the real economy, market participants are attempting to more accurately capture changes in the outlook for central banks' future monetary policies. Since the published documents are also an important tool for the central bank to communicate with the market, they are meticulously elaborated on grammatical syntax and wording, and investors are urged to read more accurately about the central bank's policy stance. Sentiment analysis on central bank documents has long been carried out, but it has been difficult to interpret the meaning of the documents accurately and to explicitly capture even the intentional change in nuance. This study attempts to evaluate the implication of the zero-shot text classification method for an unknown economic environment using the same model. We compare the tone of the statements, minutes, press conference transcripts of FOMC meetings, and the Fed officials' (chair, vice chair, and Governors) speeches. In addition, the minutes of the FOMC meetings were subjected to a phase analysis of changes in each policy stance since 1971.

## 1 Introduction

Since the monetary policies of major central banks have a broad and significant impact on financial market trends, pricing of risky assets, and spillover to the real economy, market participants are trying to better understand the changes in the future monetary policy outlook of central banks. In particular, the monetary policy of the Central Bank of the United States (Federal Reserve System, hereinafter Fed) is positioned as the most important because it influences the movement of the dollar, the key currency. One of the means by which central banks engage in dialogue with the market and conduct smooth policy management is the publication of various documents, including statements and minutes issued after policy meetings, and transcripts of speeches and congressional testimony attended by senior officials. The Federal Open Market Committee (FOMC), a meeting at which U.S. monetary policy is formulated, meets eight times a year with members of the Federal Reserve Board (FRB) and the presidents of the regional Fed banks as participants. Immediately after the meeting, a written statement is published on the website, a press conference is held by the chairperson, and transcripts of the conference are published around the next day. After approximately 3 weeks, the minutes of the FOMC meeting will be made public (They were published 3 days later until 12, 2004.). The statement is a relatively short document of about two pages that summarizes current economic perceptions, the monetary policy determined, and the names of the voters. The transcripts of the press conference consist of a transcript to be read by the chairperson at the beginning of the conference, as well as questions and answers with reporters, and are approximately 20 ~ 30 pages in volume. In some cases, information that is not included in the statement but is of interest to market participants (specific information and future prospects) is recorded. The minutes are a document that confirms the content of the economic analysis reported by the Fed economists, the process of discussion that led to the decision of the policy, and the variation of opinion among the members, and the volume is around 10~20 pages. Outside of the FOMC meetings, transcripts of speeches and interviews by FOMC participants (Fed officials) and statements in congressional testimony will be released at each meeting. Although the themes may not necessarily be related to monetary policy, the Fed officials' own views on the economy and outlook for monetary policy may be expressed, and if there is a major change from past statements, they may have an impact on the market. The statements of the members of the Fed will be published on the website, and the statements of the presidents of the regional Fed banks will be published on the website of each Fed bank. The content is often a few pages or so. Another issue that draws the attention of market participants is the Beige Book (Federal Reserve Bank Business Report). This is a report compiled by the regional Fed on the local economies of all 12 districts. It will be released on Wednesday, two weeks before the FOMC meeting, and will serve as a report to be used as a reference at the FOMC meeting. The content will be around 30 pages.

With regard to the format of the various documents, statements, minutes, and beige books are documents with a relatively standardized structure, while written records of press conferences and speeches by Fed officials are not standardized and in some cases are characterized by their colloquial tone. In terms of monetary policy implications, it is generally assumed that the transcripts of Fed officials' speeches have a head start in terms of being able to capture their perceptions ahead of FOMC meetings (However, there will be a blackout period immediately before the FOMC meeting and no Fed officials' speeches will be made.). In addition, it is considered that more information can be obtained from minutes and transcripts of conferences of the FOMC meetings with larger sentences than those with smaller sentences.

In this study, we conducted text mining using statements and minutes, transcripts of press conferences of FOMC meetings, and Fed officials' speeches, and compared their usefulness as information sources.

2 Related Work

As an application of text mining technology to the financial and economic fields, many studies of central bank documents have been conducted along with analyses of corporate accounts and news. Many studies use text mining techniques for central bank documents to score central bank sentiment, and many of them are active, aiming to predict future monetary policy, market forecasts, and economic indicators.

Ito et al. combined an expert FOMC dictionary and engagement analysis to extract sentiment by topic for the minutes of the FOMC meeting and showed that it has explanatory power for macroeconomic indicators [1]. Wang used sentence-level embedding vectors obtained with FinBERT to calculate sentiment by topic and showed that they have explanatory power for macroeconomic indicators [2].

Granziera et al. also used speech texts from FOMC members and district Fed presidents to calculate sentiment about inflation [3]. Some studies have analyzed sentiment by member using transcripts that contain all statements made by participants and are released five years after the FOMC meeting [4].

The purpose of this research is to use natural language processing techniques to read the intentions of the writer and speaker of a document as correctly as possible and to be the most correct recipient of the communication intended by the Fed. Therefore, in addition to data from 1993, when the minutes began to be published in their current format, we also analyzed the Fed's past minutes, which had been published in a similar format, going back to data from 1971 onward, and classified the characteristics of the Fed's policy tone over a long time span by interest rate hike and rate cut phases.

3 Methodology

3.1 Data and Preprocessing

We obtained the statements, minutes, transcripts of press conferences of FOMC meetings, and Fed officials' speeches published on the Fed's website[1]. In terms of FOMC minutes, paragraphs unrelated to the stance, such as the names of participants and explanations of the specific content of the policies formulated, were excluded from analysis. However, all paragraphs were covered in the old format where paragraph structure could not be obtained. The transcripts of the press conference of FOMC meetings will be published only in the manuscript portion read out by the chairperson on the day of the meeting, and the following day in the form that includes the question and answer portion by the reporters, but this time, those that include the question and answer portion were included in the analysis. However, in the question and answer section, the reporter's questions were deleted and only the chairperson's response was extracted.

The acquired documents were divided into sentence units after pre-processing such as the removal of footers and annotations.

3.2 Topic classification by the textual entailment model

Topic classification is performed using an entailment model on a zero-shot basis [6]. In this study, we used a publicly available learned model. When selecting models, members with domain knowledge compared the output of several models and selected the one that best suited their senses [2]. The entailment model returns whether the sentence to be judged contains the meaning of the hypothetical sentence with a score of 0 ~ 1. The closer the score is to 1, the more likely it is that the hypothetical statement is implied. A value closer to zero means it is inconsistent with the hypothetical statement. Since it is a zero-shot model, hypothetical statements can be freely set. In this study, we classified topics by using a hypothetical statement in the form of "This sentence is related to the topic of {}." and by setting the topic to be judged in parentheses.

---

[1] https://www.federalreserve.gov/default.htm

We set three topics, namely "Inflation", "Job Gain" and "Economic Growth" and judged each sentence. Since more than one topic may be included in a sentence, it is determined whether a sentence belongs to each topic. Since the model returns an entailment score, we set a threshold and judge that the topic belongs to the topic if the score is equal to or higher than the threshold. The threshold was set at 0.9 by checking the entailment scores of several sentences extracted as samples by members with domain knowledge.

3.3 Setting Hypothetical Statements by Category

To make an implication determination of nuances in each category, a categorical hypothesis statement is created. Hypothesis statements were set by the following process.

First, we extracted wording about the three categories of "inflation," "job gain," and "economic growth" on a keyword basis from past FOMC statements released by the Fed. Frequently used expressions were selected from the extracted expressions for each category and set as hypothetical statements. Balance of directions was also taken into consideration when selecting expressions. For example, on "Inflation": we extracted in a balanced manner "declined", "diminished", "edged down" as the expression for the downward direction, "increased", "moved up", "elevated" as the expression for the upward direction.

3.4 Entailment judgment for each category

Entailment judgment is performed for each category based on the set category-specific hypothetical statements. The target sentences for each category of entailment judgment were as follows based on the results of the 3.2 topic classification. We analyzed sentences identified as belonging to the "inflation" topic for the "inflation" category, the "Job Gain" topic for the "Job Gain" category, and the "Economic Growth" topic for the "Economic Growth" category. We used the same model as the 3.2 topic classification to determine the need for each category. Entailment judgment is performed on the categorical hypothetical sentences set in 3.3, and the number of sentences whose implication score was at or above the threshold (0.9) is counted and tabulated for each document. In each category, a stance score ($S_c(d)$) is calculated by counting the number of sentences implied in each direction of expression, taking the difference, and dividing by the total number of sentences implied in both directions of expression.

$$S_c(d) = \left( \sum_{e \in E_{c,p}} C_e(d) - \sum_{e \in E_{c,n}} C_e(d) \right) \Big/ \sum_{e \in E_{c,p}, E_{c,n}} C_e(d)$$

where $E_{c,p}$ is the set of upward expressions in the category $c$, $E_{c,n}$ is the set of downward expressions in the category $c$, and $C_e(d)$ is the number of sentences in the document $d$ judged to contain the expression $e$.

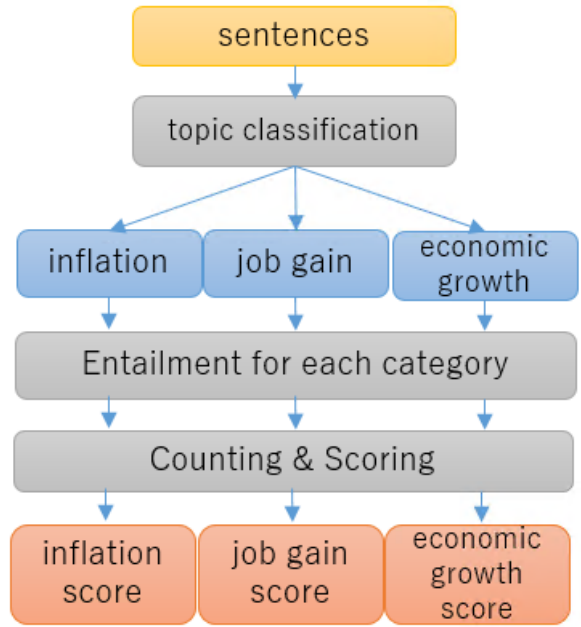

Figure 1: Analysis Flow

4 Results

4.1 Data

Documents of FOMC statements, minutes, transcripts of press conferences, and Fed officials' speeches were obtained from the FRB website and evaluated. Only minutes were obtained from February 1971, and the rest from December 2018. In addition, because it is not possible to verify the changes in policy since February 2023, which is the time point when this study was conducted, the analysis was conducted by dividing the period up to March 2022.

---
[2] https://huggingface.co/facebook/bart-large-mnli

4.2 Analysis from December 2018 to March 2022

Using data from FOMC statements, minutes, transcripts of press conferences, and Fed officials' speeches from the December 2018 meeting to the March 2022 meeting, we conducted text mining for the phases after 2021 (At a time when inflation is rising, the Fed is shifting from an accommodative policy of zero interest rates and quantitative easing to a more restrictive policy of tapering, policy rate hikes, and B/S contractions) and compared each data source. For comparison, we limited the data from December 2018, the point in time when the Chair changed the practice of holding a press conference after each FOMC meeting.

Figure 2 plots the stance scores for inflation. The score is calculated by subtracting the percentage of sentences judged to imply a hypothetical expression in the direction of inflation decline ("declined", "diminished", "moved lower", "edged down" etc.) from the percentage of sentences judged to imply a hypothetical expression in the direction of inflation rise ("increased", "picked up", "moved up", "elevated" etc.).

With regard to the FOMC statement, there was a perception that inflation was on a downward trend until the March 2021 meeting, and the score seemed to shift from the April 2021 meeting to a perception that upward pressure on inflation was intensifying. Later, the same trend continued.

Next, with regard to the score using the FOMC minutes as the data source, it was from the September 2020 meeting that the score shifted from the perception of low inflation to near neutral, and from the April 2021 meeting, as in the FOMC statement, that the perception shifted to the perception that upward pressure on the inflation rate was intensifying. Later, the same trend continued. However, it should be noted that the FOMC minutes will be published three weeks after the meeting.

Third, looking at the time series of the scores when the transcripts of the FOMC press conference are used as the data source, as in the case of the FOMC minutes, the perception shifted to the fact that the scores were gradually returning to near neutral from the September 2020 meeting and that upward pressure on the inflation rate was increasing at the March 2021 meeting. This was the earliest turnaround in comparison to the statements and the minutes of the FOMC meetings. Although there was no significant change in the inflation perception in the FOMC statement at the March 2021 meeting, the results suggest that the chairperson's post-meeting press conference may have shown his awareness of the growing inflationary pressure. The same trend continued thereafter. Given that the FOMC minutes of the meeting are released three weeks after the meeting, while the transcripts of the press conference are released the following day, we believe that this was the earliest conversion among the three data sources.

Finally, it is about the case of Fed officials' speeches as a source. For comparison with other documents at the same time, the average of the five Fed officials' speeches made immediately before the FOMC is included. As in the FOMC minutes, the score gradually turned upward from September 2020, confirming a sharp increase in April 2021.

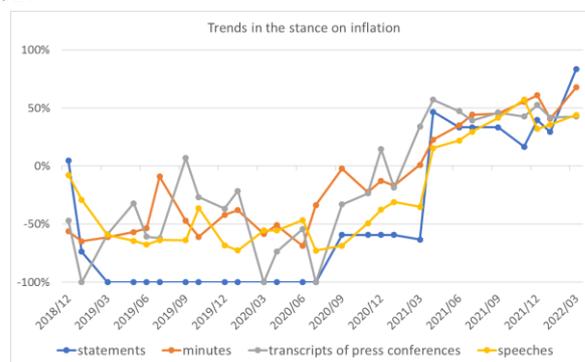

Figure 2: Trends in the stance on inflation

We then confirm the results for a similar analysis of job gain shown in Figure 3. With regard to the transition to an accommodative monetary policy following the shutdown in response to the global spread of the COVID19 since March 2020, a sharp decline in the score was confirmed in April 2020 in all sources of information, partly due to the emergency rate cut, and the perception shifted to that the job market is accommodative. Since then, press conferences and the minutes of the FOMC meetings have gradually moved back toward neutral, but only the FOMC statement sharply shifted to a perception of tight job market conditions at the April 2021 meeting.

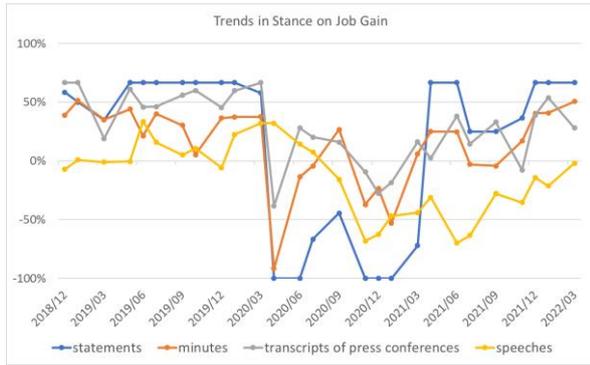

Figure 3: Trends in Stance on Job Gain

Finally, economic growth scores are shown in Figure 4. In the FOMC statements and the FOMC minutes of the February 2019 meeting, it was confirmed that economic growth had turned to a decelerating direction, leading to a rate cut in the policy rate after July 2019. Around March 2020, there was no significant difference in the timing of the score decline due to differences in data sources, partly due to the sharp economic slowdown after the shutdown. As for the subsequent progress, the scores based on the minutes, transcripts of press conferences of FOMC meetings and Fed officials' speeches returned to the same level of score around the summer of 2020 as before March 2020, while the scores for FOMC statements returned to a neutral level in the spring of 2021.

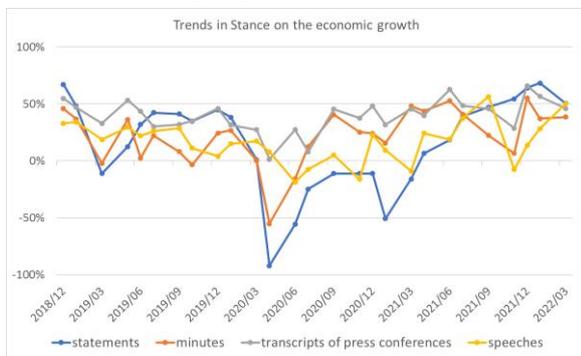

Figure 4: Trends in Stance on the economic growth

4.3 Analysis using FOMC minutes from February 1993 to March 2020

Table 1 uses the minutes of the FOMC meetings from February 1993 to March 2022 as a data source to generate scores for communication in the same method as in 4.2, and compares each aspect of the policy stance.

First, with regard to the perception of inflation, the Fed's communication, on average over the entire period, was also perceived to be under inflationary upward pressure, given that the inflation rate (annualized core PCE) has settled at a low level, with the rate remaining between 1% and 3% until 2021, since 1993. On the other hand, the scores were higher than the overall period average for periods in which the policy rate was raised, and lower than the overall period average for periods in which the rate was lowered and for periods in which the rate was zero. We believe that this result is consistent with the idea that the policy rate is changed as a means to achieve the inflation target or stabilize prices.

Next, job gain recognition is also described in Table 1. The scores were higher than the overall period average for periods in which the policy rate was raised, and lower than the overall period average for periods in which the rate was lowered and for periods in which the rate was zero. This result is consistent with the idea that the policy rate is changed as a means to achieve maximum employment, and we believe that the Fed's communication is read correctly.

Finally, a similar trend was observed for scores related to economic growth. The average over the entire period gave a positive score, i.e., a perception that the economy was strong, and a higher score than the average over the entire period during the period of interest rate increases. On the other hand, the average score was negative in the case of a rate cut, suggesting that the point at which the Fed perceived the economy to be in recession and the point at which the policy rate was actually cut were generally consistent.

Table 1: Periodic averages of scores using FOMC minutes

|  | (1) | (2) | (3) | (4) |
|---|---|---|---|---|
| Inflation | -0.23 | +0.00 | -0.40 | -0.34 |
| Job Gain | -0.08 | +0.26 | -0.45 | -0.24 |
| Economic growth | +0.22 | +0.44 | -0.16 | +0.20 |

*(1) the entire period, (2) the period of FF rate hikes, (3) the period of FF rate cuts, and (4) the period of zero interest rates (2009/1 to 2015/10, 2020/4 to 2022/1).

*(2) when the rate is raised every meeting, the meeting that kept the rate unchanged was included in the period for raising the rate.

*In the Welch t-test using the alternative hypothesis of "(2) the average of the interest rate hike phases > (3) the average of the interest rate cut phases", inflation, job gain, and business conditions were all determined to be significantly different at a significance level of 1%.

4.4 Analysis using FOMC minutes from February 1971 to December 1992

Before December 1992, there were no FOMC minutes in their current form, but there were documents similar to those named "Minutes of Action". Table 2 uses the

"Minutes of Action" at the FOMC meetings from February 1971 to December 1992 as a data source, and compares each aspect of the policy stance in the same way as in 4.3. Although the score levels were different, a similar trend to 4.3 was confirmed. Comparing the overall period averages for Tables 1 and 2, we found that Table 2 had higher scores for inflation recognition, lower scores for job gain recognition, and identical scores for economic growth. This is consistent with the magnitude of the average inflation rate (core PCE) and unemployment rate for each period. In addition, in Table 2, the tendency for the average to be higher in all three categories of inflation, job gain, and economic growth during the rate hike phase than during the rate cut phase is the same as in Table 1, suggesting that the Fed's response stance to inflation and job gain data over the long term of the past 50 years is consistent and can be quantified using the model in this study.

Table 2: Periodic averages of scores using FOMC "Minutes of Action"

|  | ① | ② | ③ |
|---|---|---|---|
| Inflation | -0.05 | +0.17 | -0.34 |
| Job Gain | -0.31 | -0.13 | -0.63 |
| Economic Growth | +0.22 | +0.30 | +0.08 |

*(1) the entire period, (2) the period of FF rate hikes, and (3) the period of FF rate cuts
*In the Welch t-test using the alternative hypothesis of "(2) the average of the interest rate hike phases > (3) the average of the interest rate cut phases", inflation, job gain, and business conditions were all determined to be significantly different at a significance level of 1%.

5 Summary

In this study, we attempted to read the changes in the Fed's communication using the textual entailment model based on zero-shot text classification. Because our model uses zero-shot classification, it is possible to handle other subjects without additional learning.

Using data from December 2018 to March 2022, we attempted to compare each document published by the Fed. For the FOMC statements, the scores generated by the model used in this study showed a slightly binary shift in scores, while a more gradual change in stance was observed for minutes and transcripts of press conferences of the FOMC meetings. In addition, the average of the past five Fed officials' speeches has remained near a neutral level, and data pre-processing may be considered as a future issue.

Next, we conducted a phase analysis using the minutes of the FOMC for the long-term period from February 1971 to March 2022, and found that the Fed's communication on inflation, job gain, and economic growth was consistent with actual policy changes, which could be interpreted using the model in this study.

Acknowledgements

This research was inspired by the FY2022 first half report of the University of Tokyo's Data Science School. We would like to thank all the students who took on the task with sincerity and came up with great ideas, as well as all the TAs and Professors.

Points of Attention

The contents and views of this paper belong to the author personally and are not the official views of the company to which he belongs.